\documentclass[10pt,twocolumn,letterpaper]{article}

\usepackage{cvpr}
\usepackage{times}
\usepackage{epsfig}
\usepackage{graphicx}
\usepackage{amsmath}
\usepackage{amssymb}

\usepackage{hhline}
\usepackage[inline]{enumitem}
\usepackage{multirow}
\usepackage[normalem]{ulem}
\usepackage[dvipsnames]{xcolor}
\usepackage[pagebackref=true,breaklinks=true,letterpaper=true,colorlinks,bookmarks=false, citecolor=ForestGreen]{hyperref}

\usepackage[square,numbers,sort&compress]{natbib}
\usepackage{algpseudocode}
\usepackage[font=footnotesize,labelfont=bf]{caption}
\usepackage{array}
\usepackage{multirow}
\usepackage{booktabs}
\usepackage{algorithm}
\usepackage{subcaption}
\usepackage[normalem]{ulem}
\usepackage{xparse}
\usepackage{pifont}
\usepackage{bm}

\usepackage{listings}
\usepackage{mwe} \usepackage{makecell}
\usepackage{color, colortbl}
\usepackage{cleveref}

\crefname{section}{\S}{\S\S}
\crefname{subsection}{\S}{\S\S}
\crefformat{table}{Table~#2#1#3}
\crefformat{figure}{Fig~#2#1#3}
\crefformat{equation}{Eqn~#2#1#3}

\usepackage[scaled=0.85]{DejaVuSansMono}

\newcommand{\algorithmName}{PIRL\xspace}
\newcommand{\algorithmShort}{PIRL\xspace}
\newcommand{\jigsawNoSp}{Jigsaw}
\newcommand{\rotationNoSp}{Rotation}
\newcommand{\jigsaw}{\jigsawNoSp \xspace}
\newcommand{\rotation}{\rotationNoSp \xspace}
\newcommand{\CPCLarge}{CPC-Big \xspace}
\newcommand{\CPCLargest}{CPC-Huge \xspace}
\newcommand{\CPCLargestPlus}{CPC-Largest \xspace}

\newcommand{\resnetfifty}{ResNet-50\xspace}
\newcommand{\resnetfiftyShort}{R-50\xspace}
\newcommand{\resnetfiftyShortVtwo}{R-50v2\xspace}
\newcommand{\resnethundredShort}{R-101\xspace}
\newcommand{\resnetonefiftyShort}{R-152\xspace}
\newcommand{\resnetCPCLargestShort}{R-170\xspace}

\newcommand{\convone}{\mytexttt{conv1}}
\newcommand{\resfour}{\mytexttt{res4}}
\newcommand{\resfive}{\mytexttt{res5}}

\newcommand{\ImNetDataset}{ImageNet\xspace}
\newcommand{\ImNet}{ImageNet\xspace}

\newcommand{\YFCCone}{YFCC-1M\xspace}
\newcommand{\YFCC}{YFCC\xspace}
\newcommand{\iNat}{iNat.\xspace}
\newcommand{\VOCseven}{VOC07\xspace}
\newcommand{\VOCseventwelve}{VOC07+12\xspace}
\newcommand{\Placestwo}{Places205\xspace}

\newcommand{\mytexttt}[1]{{\texttt{#1}}}
\newcommand{\detectrontwo}{Detectron2}

\newcommand{\makeMathAndSpace}[1]{$#1$\xspace}

\newcommand{\imageTransformNoMath}{t}
\newcommand{\imageTransformSetNoMath}{\mathcal{T}}

\newcommand{\memoryElementNoMath}{\mathbf{m}}
\newcommand{\memoryBankNoMath}{\mathcal{M}}

\newcommand{\memoryBank}{\makeMathAndSpace{\memoryBankNoMath}}

\newcommand{\imageNoMath}{\mathbf{I}}
\newcommand{\imageTransformedNoMath}{\mathbf{I}^\imageTransformNoMath}

\newcommand{\image}{\makeMathAndSpace{\imageNoMath}}
\newcommand{\imageTransformed}{\makeMathAndSpace{\imageTransformedNoMath}}

\newcommand{\calD}{\mathcal{D}}
\newcommand{\calT}{\mathcal{T}}
\newcommand{\bI}{\mathbf{I}}

\newcommand{\bv}{\mathbf{v}}

\newcommand{\tempNoMath}{\tau}

\newcommand{\numNegNoMath}{N}
\newcommand{\numNeg}{\makeMathAndSpace{\numNegNoMath}}

\newlength\savewidth\newcommand\shline{\noalign{\global\savewidth\arrayrulewidth
  \global\arrayrulewidth 1pt}\hline\noalign{\global\arrayrulewidth\savewidth}}

\newlength\thinwidth\newcommand\thinline{\noalign{\global\savewidth\arrayrulewidth
  \global\arrayrulewidth 0.5pt}\hline\noalign{\global\arrayrulewidth\savewidth}}

\definecolor{Gray}{gray}{0.92}
\newcolumntype{a}{>{\columncolor{Gray}}c}
\definecolor{LightCyan}{rgb}{0.88,1,1}
\definecolor{highlightRowColor}{gray}{0.92}

\cvprfinalcopy

\begin{document}

\title{Self-Supervised Learning of Pretext-Invariant Representations}

\author{
Ishan Misra \quad \quad
Laurens van der Maaten \\
Facebook AI Research
}

\maketitle

\begin{abstract}
The goal of self-supervised learning from images is to construct image representations that are semantically meaningful via pretext tasks that do not require semantic annotations for a large training set of images. Many pretext tasks lead to representations that are \emph{covariant} with image transformations. We argue that, instead, semantic representations ought to be \emph{invariant} under such transformations. Specifically, we develop Pretext-Invariant Representation Learning (\algorithmName, pronounced as ``pearl'') that learns invariant representations based on pretext tasks. We use \algorithmName with a commonly used pretext task that involves solving jigsaw puzzles. We find that \algorithmName substantially improves the semantic quality of the learned image representations. Our approach sets a new state-of-the-art in self-supervised learning from images on several popular benchmarks for self-supervised learning. Despite being unsupervised, \textbf{\algorithmName outperforms supervised pre-training} in learning image representations for object detection. Altogether, our results demonstrate the potential of self-supervised learning of image representations with good invariance properties.
\end{abstract}

\section{Introduction}
\label{sec:introduction}
Modern image-recognition systems learn image representations from large collections of images and corresponding semantic annotations. These annotations can be provided in the form of class labels~\cite{ILSVRC15}, hashtags~\cite{mahajan2018exploring}, bounding boxes~\cite{Everingham15,COCO}, \emph{etc.} Pre-defined semantic annotations scale poorly to the long tail of visual concepts~\cite{vanhorn2017devil}, which hampers further improvements in image recognition.

\begin{figure}[t]
    \centering
    \includegraphics[width=0.93\linewidth]{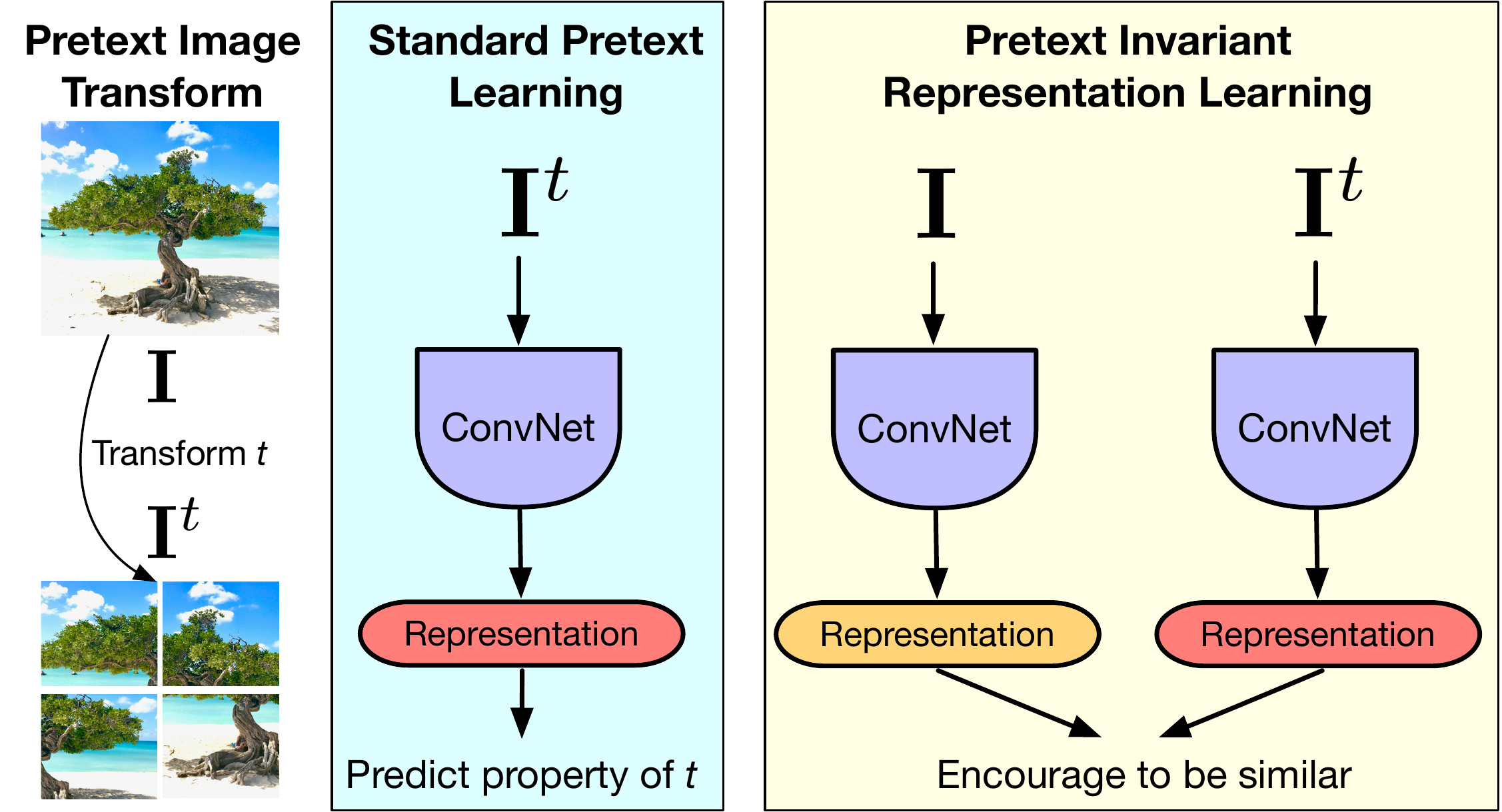}
    \vspace{-0.08in}
    \caption{\textbf{Pretext-Invariant Representation Learning (\algorithmName).} Many pretext tasks for self-supervised learning~\cite{gidaris2018unsupervised,noroozi2016unsupervised,zhang2019aet} involve transforming an image $\bI$, computing a representation of the transformed image, and predicting properties of transformation $t$ from that representation. As a result, the representation must \emph{covary} with the transformation $t$ and may not contain much semantic information. By contrast, \algorithmName learns representations that are \emph{invariant} to the transformation $t$ and retain semantic information.
    }
    \label{fig:teaser}
\end{figure}

Self-supervised learning tries to address these limitations by learning image representations from the pixels themselves without relying on pre-defined semantic annotations. Often, this is done via a \emph{pretext} task that applies a transformation to the input image and requires the learner to predict properties of the transformation from the transformed image (see Figure~\ref{fig:teaser}). Examples of image transformations used include rotations~\cite{gidaris2018unsupervised}, affine transformations~\cite{zhang2019aet,novotny2018self,rocco2017convolutional,kanazawa2016warpnet}, and jigsaw transformations~\cite{noroozi2016unsupervised}. As the pretext task involves predicting a property of the image transformation, it encourages the construction og image representations that are \emph{covariant} to the transformations. Although such covariance is beneficial for tasks such as predicting 3D correspondences~\cite{novotny2018self,kanazawa2016warpnet,rocco2017convolutional}, it is undesirable for most semantic recognition tasks. Representations ought to be \emph{invariant} under image transformations to be useful for image recognition~\cite{ji2019invariant,dosovitskiy2016discriminative} because the transformations do not alter visual semantics.

\begin{figure}
\centering
\includegraphics[width=\linewidth]{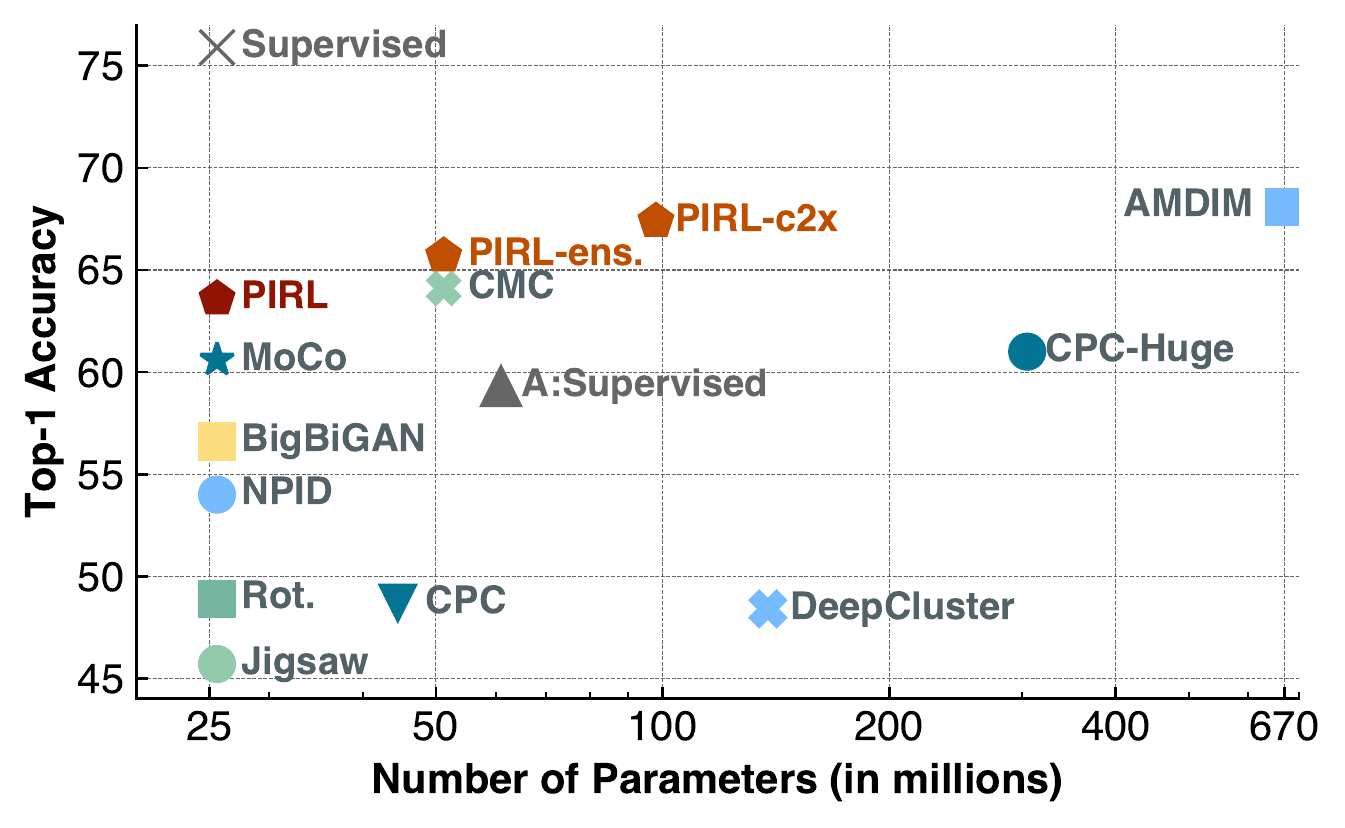}
\caption{\textbf{\ImNet classification with linear models.} Single-crop top-1 accuracy on the \ImNet validation data as a function of the number of parameters in the model that produces the representation (``A'' represents AlexNet). Pretext-Invariant Representation Learning (\algorithmName) sets a new state-of-the-art in this setting ({\color{Maroon}red marker}) and uses significantly smaller models (\resnetfifty). See Section~\ref{sec:linear_all} for more details.
}
\label{fig:imagenet_transfer}
\end{figure}

Motivated by this observation, we propose a method that learns invariant representations rather than covariant ones. Instead of predicting properties of the image transformation, Pretext-Invariant Representation Learning (\algorithmName) constructs image representations that are \emph{similar} to the representation of transformed versions of the same image and \emph{different} from the representations of other images. We adapt the ``\jigsaw'' pretext task~\cite{noroozi2016unsupervised} to work with \algorithmName and find that the resulting invariant representations perform better than their covariant counterparts across a range of vision tasks. \algorithmName substantially outperforms all prior art in self-supervised learning from ImageNet (Figure~\ref{fig:imagenet_transfer}) and from uncurated image data (Table~\ref{tab:linear_yfcc_1M}). Interestingly, \algorithmName even outperforms supervised pre-training in learning image representations suitable for object detection (Tables~\ref{tab:detection_voc07} \&~\ref{tab:detection_voc07_train}).

\section{PIRL: Pretext-Invariant \\ Representation Learning}
\label{sec:approach}

Our work focuses on pretext tasks for self-supervised learning in which a known image transformation is applied to the input image. For example, the ``\jigsaw'' task divides the image into nine patches and perturbs the image by randomly permuting the patches~\cite{noroozi2016unsupervised}. Prior work used \jigsaw as a pretext task by predicting the permutation from the perturbed input image. This requires the learner to construct a representation that is \emph{covariant} to the perturbation. The same is true for a range of other pretext tasks that have recently been studied~\cite{gidaris2018unsupervised,doersch2015unsupervised,misra2016shuffle,zhang2019aet}.
In this work, we adopt the existing \jigsaw pretext task in a way that encourages the image representations to be \emph{invariant} to the image patch perturbation. While we focus on the \jigsaw pretext task in this paper, our approach is applicable to any pretext task that involves image transformations (see Section~\ref{sec:rotation}).

\subsection{Overview of the Approach}
\label{sec:approach_overview}
Suppose we are given an image dataset, $\calD = \{\bI_1, \dots, \bI_{| \calD |}\}$ with $\bI_n \in \mathbb{R}^{H \times W \times 3}$, and a set of image transformations, $\calT$. The set $\calT$ may contain transformations such as a re-shuffling of patches in the image~\cite{noroozi2016unsupervised}, image rotations~\cite{gidaris2018unsupervised}, \etc. We aim to train a convolutional network, $\phi_\theta(\cdot)$, with parameters $\theta$ that constructs image representations $\bv_\bI = \phi_\theta(\bI)$ that are invariant to image transformations $t \in \calT$. We adopt an empirical risk minimization approach to learning the network parameters $\theta$. Specifically, we train the network by minimizing the empirical risk:
\begin{equation}
\ell_{inv}(\theta; \calD) = \mathbb{E}_{t \sim p(\calT)} \left[ \frac{1}{| \calD |} \sum_{\bI \in \calD} L\left(\bv_\bI, \bv_{\bI^t}\right) \right],
\label{eq:invariant_loss}
\end{equation}
where $p(\calT)$ is some distribution over the transformations in $\calT$, and $\bI^t$ denotes image $\bI$ after application of transformation $t$, that is, $\bI^t = t(\bI)$. The function $L(\cdot, \cdot)$ is a loss function that measures the similarity between two image representations. Minimization of this loss encourages the network $\phi_\theta(\cdot)$ to produce the same representation for image $\bI$ as for its transformed counterpart $\bI^t$, \emph{i.e.}, to make representation invariant under transformation $t$. 

We contrast our loss function to losses~\cite{gidaris2018unsupervised,doersch2015unsupervised,misra2016shuffle,noroozi2016unsupervised,zhang2019aet} that aim to learn image representations $\bv_\bI = \phi_\theta(\bI)$ that are covariant to image transformations $t \in \calT$ by minimizing:
\begin{equation}
\ell_{co}(\theta; \calD) = \mathbb{E}_{t \sim p(\calT)} \left[ \frac{1}{| \calD |} \sum_{\bI \in \calD} L_{co}\left(\bv_\bI, z(t) \right) \right],
\label{eq:covariant_loss}
\end{equation}
where $z$ is a function that measures some properties of transformation $t$. Such losses encourage network $\phi_\theta(\cdot)$ to learn image representations that contain information on transformation $t$, thereby encouraging it to maintain information that is not semantically relevant.

\par \noindent \textbf{Loss function.} We implement $\ell_{inv}(\cdot)$ using a contrastive loss function $L(\cdot, \cdot)$~\cite{hadsell2006dimensionality}. Specifically, we define a matching score, $s(\cdot, \cdot)$, that measures the similarity of two image representations and use this matching score in a noise contrastive estimator~\cite{gutmann2010noise}. In our noise contrastive estimator (NCE), each ``positive'' sample $(\bI, \bI^t)$ has \numNeg corresponding ``negative'' samples. The negative samples are obtained by computing features from other images, $\bI' \neq \bI$. The noise contrastive estimator models the probability of the binary event that $(\bI, \bI^t)$ originates from data distribution as:
\begin{equation}
h(\bv_\bI, \bv_{\bI^t}) = \frac{\exp\left(\frac{s(\bv_\bI, \bv_{\bI^t})}{\tau}\right)}{\exp\left(\frac{s(\bv_\bI, \bv_{\bI^t})}{\tau}\right) + \sum_{\bI' \in \calD_N} \exp\left(\frac{s(\bv_{\bI^t}, \bv_{\bI'})}{\tau}\right)}.
\label{eq:nce_prob}
\end{equation}
Herein, $\calD_N \subseteq \calD \setminus \{\bI\}$ is a set of $N$ negative samples that are drawn uniformly at random from dataset $\calD$ excluding image $\bI$, $\tau$ is a temperature parameter, and $s(\cdot, \cdot)$ is the cosine similarity between the representations.

\begin{figure}
\includegraphics[width=0.95\linewidth]{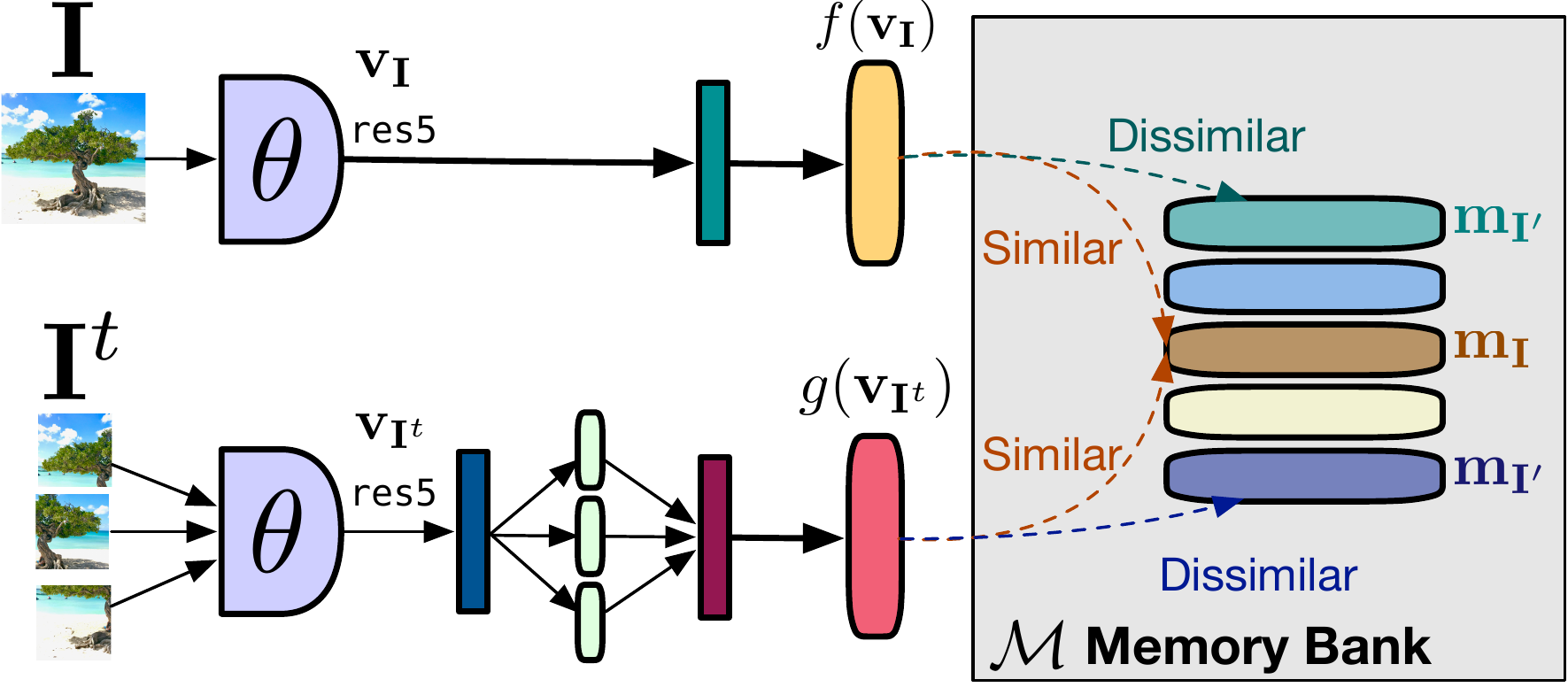}
\caption{\textbf{Overview of \algorithmName.} Pretext-Invariant Representation Learning (\algorithmName) aims to construct image representations that are invariant to the image transformations $t \in \calT$. \algorithmName encourages the representations of the image, $\bI$, and its transformed counterpart, $\bI^t$, to be similar. It achieves this by minimizing a contrastive loss (see Section~\ref{sec:approach_overview}). Following~\cite{wu2018unsupervised}, \algorithmName uses a memory bank, \memoryBank, of negative samples to be used in the contrastive learning. The memory bank contains a moving average of representations, $\memoryElementNoMath_{\bI} \in \mathcal{M}$, for all images in the dataset (see Section~\ref{sec:neg_samples}).}
\label{fig:approach}
\end{figure}

In practice, we do not use the convolutional features $\bv$ directly but apply to different ``heads'' to the features before computing the score $s(\cdot, \cdot)$. Specifically, we apply head $f(\cdot)$ on features ($\bv_\bI$) of $\bI$ and head $g(\cdot)$ on features ($\bv_{\bI^t}$) of $\bI^{t}$; see Figure~\ref{fig:approach} and Section~\ref{sec:implementation_details}. NCE then amounts to minimizing the following loss:
\begin{align}
\label{eq:nce_loss}
L_{\mathrm{NCE}}\left(\bI, \bI^t\right) = - &\log \left[ h\left(f(\bv_\bI), g(\bv_{\bI^t})\right) \right] \\
- \sum_{\bI' \in \calD_N} &\log\left[ 1 - h\left(g(\bv_\bI^t), f(\bv_{\bI'})\right) \right].
\nonumber
\end{align}
This loss encourages the representation of image $\bI$ to be similar to that of its transformed counterpart $\bI^t$, whilst also encouraging the representation of $\bI^t$ to be dissimilar to that of other images $\bI'$.

\subsection{Using a Memory Bank of Negative Samples}
\label{sec:neg_samples}
Prior work has found that it is important to use a large number of negatives in the NCE loss of Equation~\ref{eq:nce_loss}~\cite{oord2018representation,wu2018unsupervised}. In a mini-batch SGD optimizer, it is difficult to obtain a large number of negatives without increasing the batch to an infeasibly large size. To address this problem, we follow~\cite{wu2018unsupervised} and use a memory bank of ``cached'' features. Concurrent work used a similar memory-bank approach~\cite{he2019moco}.

The memory bank, \memoryBank, contains a feature representation $\memoryElementNoMath_{\bI}$ for each image $\bI$ in dataset $\calD$. The representation $\memoryElementNoMath_{\bI}$ is an exponential moving average of feature representations $f(\bv_{\bI})$ that were computed in prior epochs. This allows us to replace negative samples, $f(\bv_\bI')$, by their memory bank representations, $\memoryElementNoMath_{\bI'}$, in Equation~\ref{eq:nce_loss} without having to increase the training batch size. We emphasize that the representations that are stored in the memory bank are all computed on the original images, $\bI$, without the transformation $t$.

\par \noindent \textbf{Final loss function.} A potential issue of the loss in Equation~\ref{eq:nce_loss} is that it does not compare the representations of untransformed images $\bI$ and $\bI'$. We address this issue by using a convex combination of two NCE loss functions in $\ell_{inv}(\cdot)$:
\begin{align}
\label{eq:final_loss}
L\left(\bI, \bI^t\right) = \lambda L_{\mathrm{NCE}}(\memoryElementNoMath_{\bI}, g(\bv_{\bI^{t}})) \notag \\
+ (1 - \lambda) L_{\mathrm{NCE}}(\memoryElementNoMath_{\bI}, f(\bv_{\bI})).
\end{align}
Herein, the first term is simply the loss of Equation~\ref{eq:nce_loss} but uses memory representations $\memoryElementNoMath_\bI$ and $\memoryElementNoMath_{\bI'}$ instead of $f(\bv_\bI)$ and $f(\bv_\bI')$, respectively. The second term does two things: (1) it encourages the representation $f(\bv_\bI)$ to be similar to its memory representation $\memoryElementNoMath_\bI$, thereby dampening the parameter updates; and (2) it encourages the representations $f(\bv_\bI)$ and $f(\bv_\bI')$ to be dissimilar. We note that both the first and the second term use $\memoryElementNoMath_{\bI'}$ instead of $f(\bv_\bI')$ in Equation~\ref{eq:nce_loss}. Setting $\lambda=0$ in Equation~\ref{eq:final_loss} leads to the loss used in~\cite{wu2018unsupervised}. We study the effect of $\lambda$ on the learned representations in Section~\ref{sec:analysis}.

\subsection{Implementation Details}
\label{sec:implementation_details}

Although \algorithmName can be used with any pretext task that involves image transformations, we focus on the \jigsaw pretext task~\cite{noroozi2016unsupervised} in this paper. To demonstrate that \algorithmName is more generally applicable, we also experiment with the \rotation pretext task~\cite{gidaris2018unsupervised} and with a combination of both tasks in Section~\ref{sec:rotation}. Below, we describe the implementation details of \algorithmName with the \jigsaw pretext task.

\par \noindent \textbf{Convolutional network.} We use a \resnetfifty (\resnetfiftyShort) network architecture in our experiments~\cite{he2016deep}. The network is used to compute image representations for both \image and \imageTransformed. These representations are obtained by applying function $f(\cdot)$ or $g(\cdot)$ on features extracted from the the network.

Specifically, we compute the representation of $\imageNoMath$, $f(\bv_\bI)$, by extracting \resfive~features, average pooling, and a linear projection to obtain a $128$-dimensional representation. 

To compute the representation $g(\bv_{\bI^t})$ of a transformed image \imageTransformed, we closely follow~\cite{noroozi2016unsupervised,goyal2019scaling}. We: (1) extract nine patches from image $\bI$, (2) compute an image representation for each patch separately by extracting activations from the \resfive~layer of the \resnetfifty and average pool the activations, (3) apply a linear projection to obtain a $128$-dimensional patch representations, and (4) concatenate the patch representations in random order and apply a second linear projection on the result to obtain the final $128$-dimensional image representation, $g(\bv_{\bI^t})$. Our motivation for this design of $g(\bv_{\bI^t})$ is the desire to remain as close as possible to the covariant pretext task of~\cite{noroozi2016unsupervised,gidaris2018unsupervised,goyal2019scaling}. This allows apples-to-apples comparisons between the covariant approach and our invariant approach.

\par \noindent \textbf{Hyperparameters.} We implement the memory bank as described in~\cite{wu2018unsupervised} and use the same hyperparameters for the memory bank. Specifically, we set the temperature in Equation~\ref{eq:nce_prob} to $\tempNoMath\!=\!0.07$, and use a weight of $0.5$ to compute the exponential moving averages in the memory bank. Unless stated otherwise, we use $\lambda\!=0.5$ in Equation~\ref{eq:final_loss}.

\section{Experiments}
\label{sec:experiments}

Following common practice in self-supervised learning~\cite{zhang2017split,goyal2019scaling}, we evaluate the performance of \algorithmName in transfer-learning experiments. We perform experiments on a variety of datasets, focusing on object detection and image classification tasks. Our empirical evaluations cover: (1) a learning setting in which the parameters of the convolutional network are \emph{finetuned} during transfer, thus evaluating the network ``initialization'' obtained using self-supervised learning and (2) a learning setting in which the parameters of the network are \emph{fixed} during transfer learning, thus using the network as a feature extractor. Code reproducing the results of our experiments will be published online.  

\par \noindent \textbf{Baselines.} Our most important baseline is the \jigsaw \resnetfifty model of~\cite{goyal2019scaling}. This baseline implements the covariant counterpart of our \algorithmName approach with the \jigsaw pretext task.

We also compare \algorithmName to a range of other self-supervised methods. An important comparison is to NPID~\cite{wu2018unsupervised}. NPID is a special case of \algorithmName: setting $\lambda = 0$ in Equation~\ref{eq:final_loss} leads to the loss function of NPID. We found it is possible to improve the original implementation of NPID by using more negative samples and training for more epochs (see Section~\ref{sec:analysis}). We refer to our improved version of NPID as NPID++. Comparisons between \algorithmName and NPID++ allow us to study the effect of the pretext-invariance that \algorithmName aims to achieve, \emph{i.e.}, the effect of using $\lambda > 0$ in Equation~\ref{eq:final_loss}.

\par \noindent \textbf{Pre-training data.} To facilitate comparisons with prior work, we use the $1.28$M images from the \ImNetDataset~\cite{ILSVRC15} \mytexttt{train} split (without labels) to pre-train our models.

\par \noindent \textbf{Training details.} We train our models using mini-batch SGD using the cosine learning rate decay~\cite{loshchilov2016sgdr} scheme with an initial learning rate of $1.2\times10^{-1}$ and a final learning rate of $1.2\times10^{-4}$. We train the models for $800$ epochs using a batch size of $1,024$ images and using $N=32,000$ negative samples in Equation~\ref{eq:nce_prob}.  We do not use data-augmentation approaches such as Fast AutoAugment~\cite{lim2019fast} because they are the result of supervised-learning approaches. We provide a full overview of all hyperparameter settings that were used in the supplemental material.

\par \noindent \textbf{Transfer learning.} Prior work suggests that the hyperparameters used in transfer learning can play an important role in the evaluation pre-trained representations~\cite{zhang2017split,goyal2019scaling,kolesnikov2019revisiting}. To facilitate fair comparisons with prior work, we closely follow the transfer-learning setup described in~\cite{goyal2019scaling,zhang2017split}.

\begin{table}[!t]
\centering
\setlength{\tabcolsep}{0.2em}\resizebox{\linewidth}{!}{
\begin{tabular}{l|c|ccc|r}
\textbf{Method} & \textbf{Network} & $\boldsymbol{\mathrm{AP}^\mathrm{all}}$ & $\boldsymbol{\mathrm{AP}^{50}}$ & $\boldsymbol{\mathrm{AP}^{75}}$ & $\boldsymbol{\Delta \mathrm{AP}^{75}}$ \\
\shline

Supervised & \resnetfiftyShort & 52.6 & \textbf{81.1} & 57.4 & =0.0\\
\thinline
\rowcolor{highlightRowColor} \jigsaw~\cite{goyal2019scaling} & \resnetfiftyShort & 48.9 & 75.1 & 52.9 & -4.5\\
\rotation~\cite{goyal2019scaling} & \resnetfiftyShort & 46.3 & 72.5 & 49.3 & -8.1\\
\rowcolor{highlightRowColor} NPID++~\cite{wu2018unsupervised} & \resnetfiftyShort & 52.3 & 79.1 & 56.9 & -0.5\\
\algorithmShort \textbf{(ours)} & \resnetfiftyShort & \textbf{54.0}	& \underline{80.7} & \textbf{59.7} & \textbf{+2.3}\\

\thinline
\rowcolor{highlightRowColor} \CPCLarge~\cite{henaff2019data} & \resnethundredShort & -- & ~70.6$^*$ & -- \\
\CPCLargest~\cite{henaff2019data} & \resnetCPCLargestShort & -- & ~72.1$^*$ & -- \\
\rowcolor{highlightRowColor} MoCo~\cite{he2019moco} & \resnetfiftyShort & 55.2$^{*\dagger}$ & 81.4$^{*\dagger}$ & 61.2$^{*\dagger}$ \\
\thinline

\end{tabular}}

\caption{\textbf{Object detection on \VOCseventwelve using Faster R-CNN.} Detection AP on the \VOCseven test set after finetuning Faster R-CNN models (keeping BatchNorm fixed) with a \resnetfifty backbone pre-trained using self-supervised learning on \ImNet. Results for supervised \ImNet pre-training are presented for reference. Numbers with $^*$ are adopted from the corresponding papers. Method with $^\dagger$ finetunes BatchNorm. \algorithmName significantly outperforms supervised pre-training without extra pre-training data or changes in the network architecture. Additional results in Table~\ref{tab:detection_voc07_train}.}
\label{tab:detection_voc07}
\end{table}

\subsection{Object Detection}
\label{sec:detection}
Following prior work~\cite{goyal2019scaling,wu2018unsupervised}, we perform object-detection experiments on the the Pascal VOC dataset~\cite{Everingham15} using the VOC07+12 train split. We use the Faster R-CNN~\cite{ren2015faster} C4 object-detection model implemented in \detectrontwo~\cite{wu2019detectron2} with a \resnetfifty (\resnetfiftyShort) backbone. We pre-train the \resnetfifty using \algorithmName to initialize the detection model before \underline{finetuning} it on the VOC training data. We use the same training schedule as~\cite{goyal2019scaling} for all models finetuned on VOC and follow~\cite{goyal2019scaling,wu2019detectron2} to keep the BatchNorm parameters fixed during finetuning. We evaluate object-detection performance in terms of AP$^{\mathrm{all}}$, AP$^{50}$, and AP$^{75}$~\cite{COCO}.

The results of our detection experiments are presented in Table~\ref{tab:detection_voc07}. The results demonstrate the strong performance of \algorithmName: it outperforms all alternative self-supervised learnings in terms of all three AP measures. Compared to pre-training on the Jigsaw pretext task, \algorithmName achieves AP improvements of $\mathbf{5}$ \textbf{points}. These results underscore the importance of learning invariant (rather than covariant) image representations. \algorithmName also outperforms NPID++, which demonstrates the benefits of learning pretext invariance.

Interestingly, {\algorithmName even outperforms the supervised \ImNetDataset-pretrained model} in terms of the more conservative AP$^{\mathrm{all}}$ and AP$^{75}$ metrics. Similar to concurrent work~\cite{he2019moco}, we find that a self-supervised learner can \textbf{outperform supervised} pre-training for object detection. We emphasize that \algorithmName achieves this result using the \emph{same} backbone model, the \emph{same} number of finetuning epochs, and the exact \emph{same} pre-training data (but without the labels). This result is a substantial improvement over prior self-supervised approaches that obtain slightly worse performance than fully supervised baselines despite using orders of magnitude more curated training data~\cite{goyal2019scaling} or much larger backbone models~\cite{henaff2019data}. In Table~\ref{tab:detection_voc07_train}, we show that \algorithmName also outperforms supervised pretraining when finetuning is done on the much smaller \VOCseven\ train+val set. This suggests that \algorithmName learns image representations that are amenable to sample-efficient supervised learning.

\begin{table}[!t]
\centering
\setlength{\tabcolsep}{0.2em}\resizebox{\linewidth}{!}{
\begin{tabular}{l|c|acac}
\textbf{Method} & \textbf{Parameters} & \multicolumn{4}{c}{\textbf{Transfer Dataset}}\\
 & & {\small{\textbf{\ImNet}}} & {\small{\textbf{\VOCseven}}} & {\small{\textbf{\Placestwo}}} & {\small{\textbf{\iNat}}} \\
\shline
\multicolumn{6}{c}{\resnetfifty using evaluation setup of~\cite{goyal2019scaling}} \\
\thinline
Supervised & 25.6M & 75.9 & 87.5 & 51.5 & 45.4\\
\thinline
Colorization~\cite{goyal2019scaling} & 25.6M & 39.6 & 55.6 & 37.5 & -- \\

\rotation~\cite{gidaris2018unsupervised} & 25.6M & 48.9 & 63.9 & 41.4 & 23.0\\
NPID++~\cite{wu2018unsupervised}  & 25.6M & 59.0 & 76.6 & 46.4 & 32.4 \\
MoCo~\cite{he2019moco} & 25.6M & 60.6 & -- & -- & -- \\
\jigsaw~\cite{goyal2019scaling} & 25.6M & 45.7 & 64.5 & 41.2 & 21.3\\
\algorithmShort (\textbf{ours})& 25.6M & \textbf{63.6} & \textbf{81.1} & \textbf{49.8} & \textbf{34.1} \\

\thinline
\multicolumn{6}{c}{Different architecture or evaluation setup}\\
\thinline
NPID~\cite{wu2018unsupervised} & 25.6M & 54.0 & -- & 45.5 & -- \\
BigBiGAN~\cite{donahue2019large} & 25.6M & 56.6 & -- & -- & -- \\

AET~\cite{zhang2019aet} & 61M & 40.6 & -- & 37.1 & -- \\

DeepCluster~\cite{caron2018deep} & 61M & 39.8 & -- & 37.5 & -- \\

Rot.~\cite{kolesnikov2019revisiting} & 61M & 54.0 & -- & 45.5 & -- \\

LA~\cite{zhuang2019local} & 25.6M & ~60.2$^\dagger$ & -- & ~50.2$^\dagger$ & -- \\

CMC~\cite{tian2019contrastive} & 51M & 64.1 & -- & -- & -- \\

CPC~\cite{oord2018representation} & 44.5M & 48.7 & -- & -- & -- \\
\CPCLargest~\cite{henaff2019data} & 305M & 61.0 & -- & -- & -- \\
BigBiGAN-Big~\cite{donahue2019large} & 86M & 61.3 & -- & -- & -- \\
AMDIM~\cite{bachman2019amdim} & 670M & 68.1 & -- & 55.1 & -- \\
\thinline
\thinline

\end{tabular}}

\caption{\textbf{Image classification with linear models.} Image-classification performance on four datasets using the setup of~\cite{goyal2019scaling}. We train linear classifiers on image representations obtained by self-supervised learners that were pre-trained on \ImNet (without labels). We report the performance for the best-performing layer for each method. We measure mean average precision (mAP) on the \VOCseven dataset and top-1 accuracy on all other datasets. Numbers for \algorithmShort, NPID++, Rotation were obtained by us; the other numbers were adopted from their respective papers. Numbers with $^\dagger$ were measured using 10-crop evaluation. The best-performing self-supervised learner on each dataset is \textbf{boldfaced}.}
\label{tab:linear_all}
\end{table}

\subsection{Image Classification with Linear Models}
\label{sec:linear_all}

Next, we assess the quality of image representations by training linear classifiers on \underline{fixed} image representations. We follow the evaluation setup from~\cite{goyal2019scaling} and measure the performance of such classifiers on four image-classification datasets: \ImNetDataset~\cite{ILSVRC15}, \VOCseven~\cite{Everingham15}, \Placestwo~\cite{zhou2014learning}, and iNaturalist2018~\cite{van2018inaturalist}. These datasets involve diverse tasks such as object classification, scene recognition and fine-grained recognition. Following~\cite{goyal2019scaling}, we evaluate representations extracted from all intermediate layers of the pre-trained network, and report the image-classification results for the best-performing layer in Table~\ref{tab:linear_all}.

\par \noindent \textbf{\ImNet results.} The results on \ImNet highlight the benefits of learning invariant features: \algorithmShort improves recognition accuracies by over $15\%$ compared to its covariant counterpart, \jigsaw. \algorithmShort achieves the \textbf{highest single-crop top-1} accuracy of all self-supervised learners that use a single \resnetfifty model.

The benefits of pretext invariance are further highlighted by comparing \algorithmName with NPID. Our re-implementation of NPID (called NPID++) substantially outperforms the results reported in~\cite{wu2018unsupervised}. Specifically, NPID++ achieves a single-crop top-1 accuracy of $59\%$, which is higher or on par with existing work that uses a single \resnetfifty. Yet, \algorithmShort substantially outperforms NPID++. We note that \algorithmShort also outperforms concurrent work~\cite{he2019moco} in this setting.

Akin to prior approaches, the performance of \algorithmName improves with network size.
For example, CMC~\cite{tian2019contrastive} uses a combination of two \resnetfifty models and trains the linear classifier for longer to obtain $64.1\%$ accuracy. We performed an experiment in which we did the same for \algorithmShort, and obtained a top-1 accuracy of $65.7\%$; see ``\algorithmShort-ens.'' in Figure~\ref{fig:imagenet_transfer}. To compare \algorithmName with larger models, we also performed experiments in which we followed~\cite{kolesnikov2019revisiting,zhai60s4l} by doubling the number of channels in \resnetfifty; see ``\algorithmShort-c2x'' in Figure~\ref{fig:imagenet_transfer}. \algorithmShort-c2x achieves a top-1 accuracy of $67.4\%$, which is close to the accuracy obtained by AMDIM~\cite{bachman2019amdim} with a model that has $6\times$ more parameters.

Altogether, the results in Figure~\ref{fig:imagenet_transfer} demonstrate that \algorithmName outperforms all prior self-supervised learners on \ImNet in terms of the trade-off between model accuracy and size. Indeed, \algorithmName even outperforms most self-supervised learners that use much larger models~\cite{oord2018representation,henaff2019data}.

\par \noindent \textbf{Results on other datasets.} The results on the other image-classification datasets in Table~\ref{tab:linear_all} are in line with the results on \ImNet: \algorithmShort substantially outperforms its covariant counterpart (\jigsaw). The performance of \algorithmName is within 2\% of fully supervised representations on \Placestwo, and improves the previous best results of~\cite{goyal2019scaling} on \VOCseven by more than 16 AP points. On the challenging iNaturalist dataset, which has over $8,000$ classes, we obtain a gain of 11\% in top-1 accuracy over the prior best result~\cite{gidaris2018unsupervised}. We observe that the NPID++ baseline performs well on these three datasets but is consistently outperformed by \algorithmShort. Indeed, \textbf{\algorithmShort sets a new state-of-the-art} for self-supervised representations in this learning setting on the \VOCseven, \Placestwo, and iNaturalist datasets.

\subsection{Semi-Supervised Image Classification}
\label{sec:semi_supervised}

We perform semi-supervised image classification experiments on \ImNet following the experimental setup of~\cite{henaff2019data,wu2018unsupervised,zhai60s4l}. Specifically, we randomly select $1\%$ and $10\%$ of the \ImNet training data (with labels). We \underline{finetune} our models on these training-data subsets following the procedure of~\cite{wu2018unsupervised}. Table~\ref{tab:semi_supervised} reports the top-5 accuracy of the resulting models on the \ImNet validation set.

The results further highlight the quality of the image representations learned by \algorithmName: finetuning the models on just $1\%$ ($\sim$13,000) labeled images leads to a top-5 accuracy of $57\%$. \algorithmName performs at least as well as S$^{4}$L~\cite{zhai60s4l} and better than VAT~\cite{grandvalet2005semi}, which are both methods specifically designed for semi-supervised learning. In line with earlier results, \algorithmName also outperforms \jigsaw and NPID++.

\begin{table}[!t]
\centering
\setlength{\tabcolsep}{0.4em}\resizebox{\linewidth}{!}{
\begin{tabular}{l|l|ac}
& Data fraction $\rightarrow$ & ~~~1\%~~~ & 10\% \\
\textbf{Method} & \textbf{Backbone} & \multicolumn{2}{c}{\textbf{Top-5 Accuracy}} \\
\shline
Random initialization~\cite{wu2018unsupervised} & \resnetfiftyShort & 22.0 & 59.0 \\
NPID~\cite{wu2018unsupervised} & \resnetfiftyShort & 39.2 & 77.4 \\
\jigsaw~\cite{goyal2019scaling} & \resnetfiftyShort & 45.3 & 79.3 \\
NPID++~\cite{wu2018unsupervised} & \resnetfiftyShort & 52.6 & 81.5 \\

VAT + Ent Min.~\cite{miyato2018virtual,grandvalet2005semi} & \resnetfiftyShortVtwo & 47.0 & 83.4 \\
S$^{4}$L Exemplar~\cite{zhai60s4l} & \resnetfiftyShortVtwo & 47.0 & 83.7 \\
S$^{4}$L Rotation~\cite{zhai60s4l} & \resnetfiftyShortVtwo & 53.4 & \textbf{83.8} \\
\algorithmShort (\textbf{ours}) & \resnetfiftyShort & \textbf{57.2} & \textbf{83.8} \\
\thinline
Colorization~\cite{larsson2017colorization} & \resnetonefiftyShort & 29.8 & 62.0 \\
\CPCLargestPlus~\cite{henaff2019data} & \resnetCPCLargestShort and R-11 & 64.0 & 84.9 \\
\thinline

\end{tabular}}
\vspace{-0.1in}
\caption{\textbf{Semi-supervised learning on \ImNet.} Single-crop top-5 accuracy on the \ImNet\ validation set of self-supervised models that are finetuned on 1\% and 10\% of the \ImNet training data, following~\cite{wu2018unsupervised}. All numbers except for \jigsaw, NPID++ and \algorithmShort are adopted from the corresponding papers. Best performance is \textbf{boldfaced}.}
\label{tab:semi_supervised}
\end{table}

\subsection{Pre-Training on Uncurated Image Data}
\label{sec:yfcc}

Most representation learning methods are sensitive to the data distribution used during pre-training~\cite{joulin2016learning,sun2017revisiting,mahajan2018exploring,goyal2019scaling}. To study how much changes in the data distribution impact \algorithmShort, we pre-train models on uncurated images from the unlabeled \YFCC dataset~\cite{thomee2015yfcc100m}. Following~\cite{caron2019deep,goyal2019scaling}, we randomly select a subset of 1 million images (\YFCCone) from the 100 million images in \YFCC. We pre-train  \algorithmName \resnetfifty networks on \YFCCone using the same procedure that was used for \ImNet pre-training. We evaluate using the setup in Section~\ref{sec:linear_all} by training linear classifiers on \underline{fixed} image representations.

Table~\ref{tab:linear_yfcc_1M} reports the top-1 accuracy of the resulting classifiers. In line with prior results, \algorithmShort outperforms competing self-supervised learners. In fact, \algorithmShort even outperforms \jigsaw and DeeperCluster models that were trained on $100\times$ more data from the same distribution. Comparing pre-training on \ImNet (Table~\ref{tab:linear_all}) with pre-training \YFCCone (Table~\ref{tab:linear_yfcc_1M}) leads to a mixed set of observations. On \ImNet classification, pre-training (without labels) on \ImNet works substantially better than pre-training on \YFCCone. In line with prior work~\cite{joulin2016learning,goyal2019scaling}, however, pre-training on \YFCCone leads to better representations for image classification on the \Placestwo dataset.

\begin{table}[!t]
\centering
\setlength{\tabcolsep}{0.4em}\resizebox{\linewidth}{!}{
\begin{tabular}{l|l|caca}
\textbf{Method} & \textbf{Dataset} & \multicolumn{4}{c}{\textbf{Transfer Dataset}}\\
 & & {\small{\textbf{\ImNet}}} & {\small{\textbf{\VOCseven}}} & {\small{\textbf{\Placestwo}}} & {\small{\textbf{\iNat}}} \\

\shline

\jigsaw~\cite{goyal2019scaling} & YFCC1M & -- & 64.0 & 42.1 & -- \\
DeepCluster~\cite{caron2018deep,caron2019deep} & YFCC1M & 34.1 & 63.9 & 35.4 & -- \\

\algorithmName (\textbf{ours}) & YFCC1M & \textbf{57.8} & \textbf{78.8} & \textbf{51.0} & \textbf{29.7} \\

\thinline
\jigsaw~\cite{goyal2019scaling} & YFCC100M & 48.3 & 71.0 & 44.8 & -- \\
DeeperCluster~\cite{caron2019deep} & YFCC100M & 45.6 & 73.0 & 42.1 & -- \\
\thinline
\end{tabular}
}
\vspace{-0.1in}
\caption{\textbf{Pre-training on uncurated YFCC images.} Top-1 accuracy or mAP (for \VOCseven) of linear image classifiers for four image-classification tasks, using various image representations. All numbers (except those for \algorithmShort) are adopted from the corresponding papers. Deep(er)Cluster uses VGG-16 rather than \resnetfifty. The best performance on each dataset is \textbf{boldfaced}. \underline{Top:} Representations obtained by training \resnetfifty models on a randomly selected subset of one million images. \underline{Bottom:} Representations learned from about 100 million YFCC images.}
\label{tab:linear_yfcc_1M}
\end{table}

\section{Analysis}
\label{sec:analysis}

We performed a set of experiments aimed at better understanding the properties of \algorithmName. To make it feasible to train the larger number of models needed for these experiments, we train the models we study in this section for fewer epochs ($400$) and with fewer negatives ($N=4,096$) than in Section~\ref{sec:experiments}. As a result, we obtain lower absolute performances. Apart from that, we did not change the experimental setup or any of the other hyperparameters. Throughout the section, we use the evaluation setup from Section~\ref{sec:linear_all} that trains linear classifiers on \underline{fixed} image representations to measure the quality of image representations.

\begin{figure}[!t]
\centering
\includegraphics[width=\linewidth]{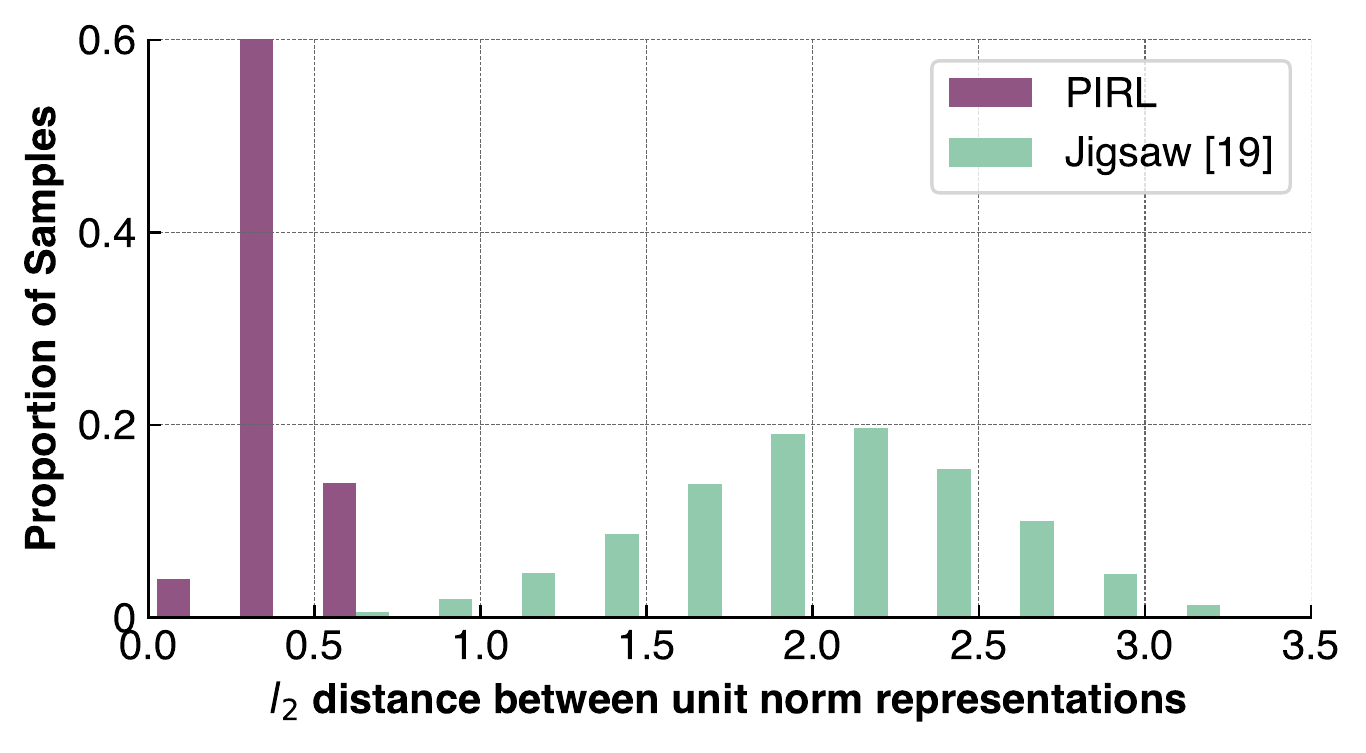}
\caption{\textbf{Invariance of \algorithmName representations.} Distribution of $l_2$ distances between unit-norm image representations, $f(\bv_\bI) / \lVert f(\bv_\bI) \rVert^2$, and unit-norm representations of the transformed image, $g(\bv_{\bI^t}) / \lVert g(\bv_{\bI^t}) \rVert^2$. Distance distributions are shown for \algorithmName and \jigsaw representations.} \label{fig:distance_histogram}
\end{figure}

\subsection{Analyzing \algorithmName Representations}

\paragraph{Does \algorithmName learn invariant representations?}~\\
\algorithmName was designed to learn representations that are invariant to image transformation $t \in \calT$. We analyzed whether the learned representations actually have the desired invariance properties. Specifically, we normalize the representations to have unit norm and compute $l_2$ distances between the (normalized) representation of image, $f(\bv_\bI)$, and the (normalized) representation its transformed version, $g(\bv_{\bI^t})$. We repeat this for all transforms $t \in \imageTransformSetNoMath$ and for a large set of images. We plot histograms of the distances thus obtained in Figure~\ref{fig:distance_histogram}. The figure shows that, for \algorithmName, an image representation and the representation of a transformed version of that image are generally similar. This suggests that \algorithmName has learned representations that are invariant to the transformations. By contrast, the distances between \jigsaw representations have a much larger mean and variance, which suggests that \jigsaw representations covary with the image transformations that were applied.

\begin{figure}[!t]
\centering
\includegraphics[width=\linewidth]{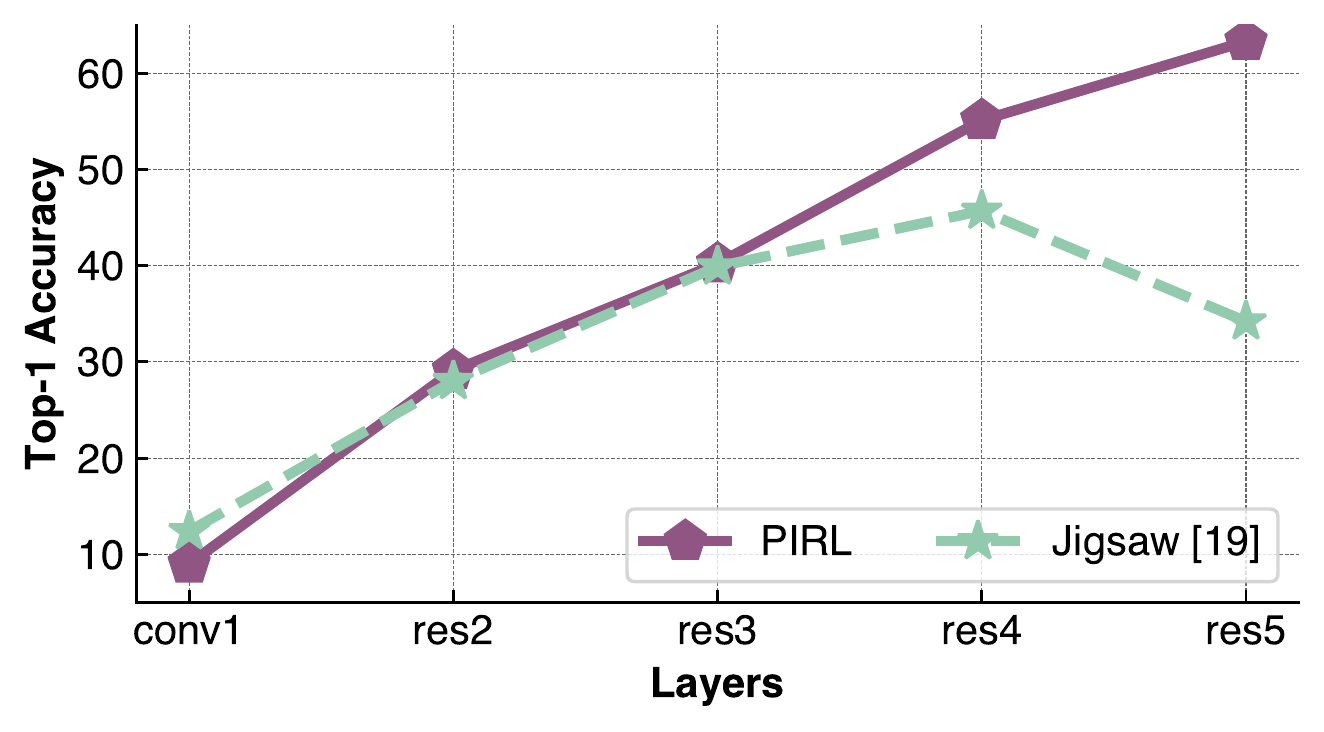}
\caption{\textbf{Quality of \algorithmName representations per layer.} Top-1 accuracy of linear models trained to predict \ImNet classes based on representations extracted from various layers in \resnetfifty trained using \algorithmName and \jigsaw.}
\vspace{-0.1in}
\label{fig:per_layer_performance}
\end{figure}

\paragraph{Which layer produces the best representations?}~\\
All prior experiments used \algorithmName representations that were extracted from the \resfive\ layer and \jigsaw representations that were extracted from the \resfour\ layer (which work better for \jigsaw).
Figure~\ref{fig:per_layer_performance} studies the quality of representations in earlier layers of the convolutional networks. The figure reveals that the quality of \jigsaw representations improves from the \convone\ to the \resfour\ layer but that their quality sharply decreases in the \resfive\ layer. We surmise this happens because the \resfive\ representations in the last layer of the network covary with the image transformation $t$ and are not encouraged to contain semantic information. By contrast, \algorithmName representations are invariant to image transformations, which allows them to focus on modeling semantic information. As a result, the best image representations are extracted from the \resfive\ layer of \algorithmName-trained networks.

\begin{figure}[!h]
\centering
\includegraphics[width=\linewidth]{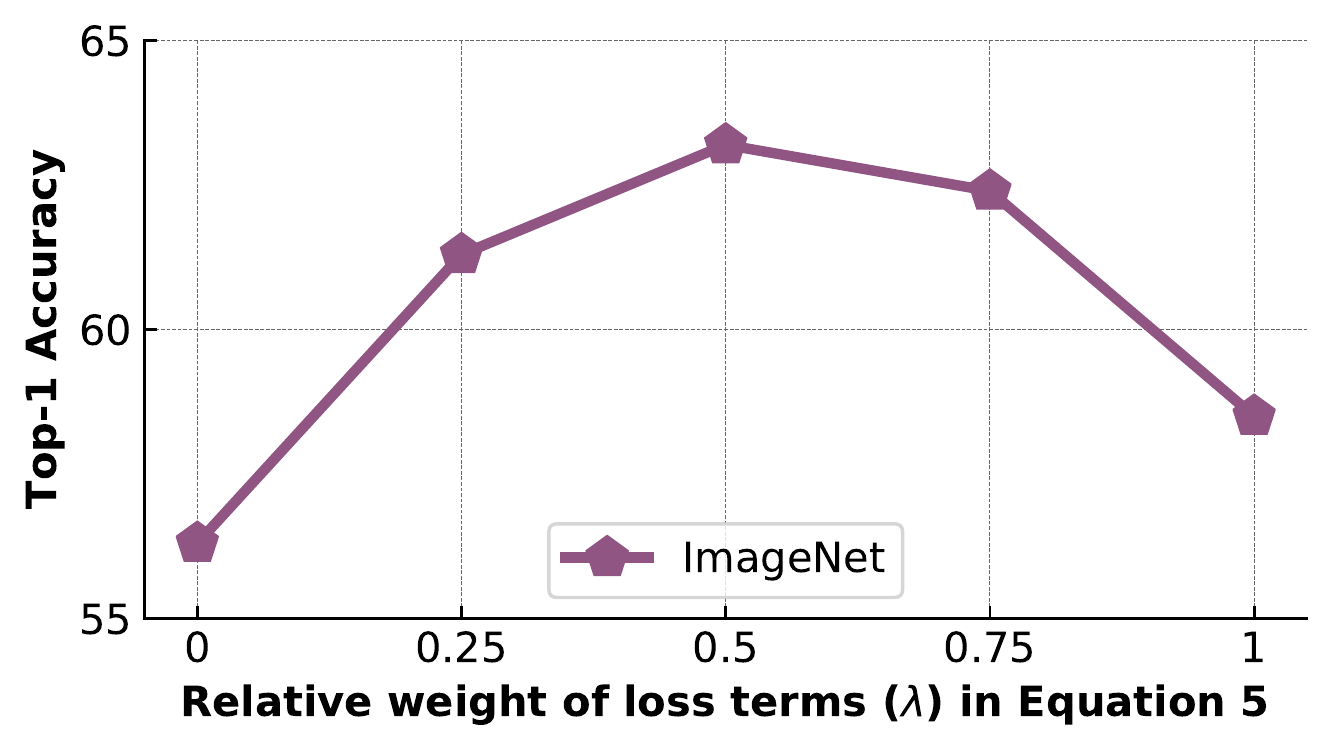}
\caption{\textbf{Effect of varying the trade-off parameter $\boldsymbol{\lambda}$.} Top-1 accuracy of linear classifiers trained to predict \ImNet classes from \algorithmName representations as a function of hyperparameter $\lambda$ in Equation~\ref{eq:final_loss}.} \label{fig:relative_loss_weight}
\end{figure}

\subsection{Analyzing the \algorithmName Loss Function}

\paragraph{What is the effect of $\lambda$ in the \algorithmName loss function?}~\\
The \algorithmName loss function in Equation~\ref{eq:final_loss} contains a hyperparameter $\lambda$ that trades off between two NCE losses. All prior experiments were performed with $\lambda\!=\!0.5$. NPID(++)~\cite{wu2018unsupervised} is a special case of \algorithmName in which $\lambda\!=\!0$, effectively removing the pretext-invariance term from the loss. At $\lambda\!=\!1$, the network does not compare untransformed images at training time and updates to the memory bank $\memoryElementNoMath_{\bI}$ are not dampened.

We study the effect of $\lambda$ on the quality of \algorithmName representations. As before, we measure representation quality by the top-1 accuracy of linear classifiers operating on \underline{fixed} \ImNet representations. Figure~\ref{fig:relative_loss_weight} shows the results of these experiments. The results show that the performance of \algorithmName is quite sensitive to the setting of $\lambda$, and that the best performance if obtained by setting $\lambda\!=\!0.5$.

\begin{figure}
\centering
\includegraphics[width=\linewidth]{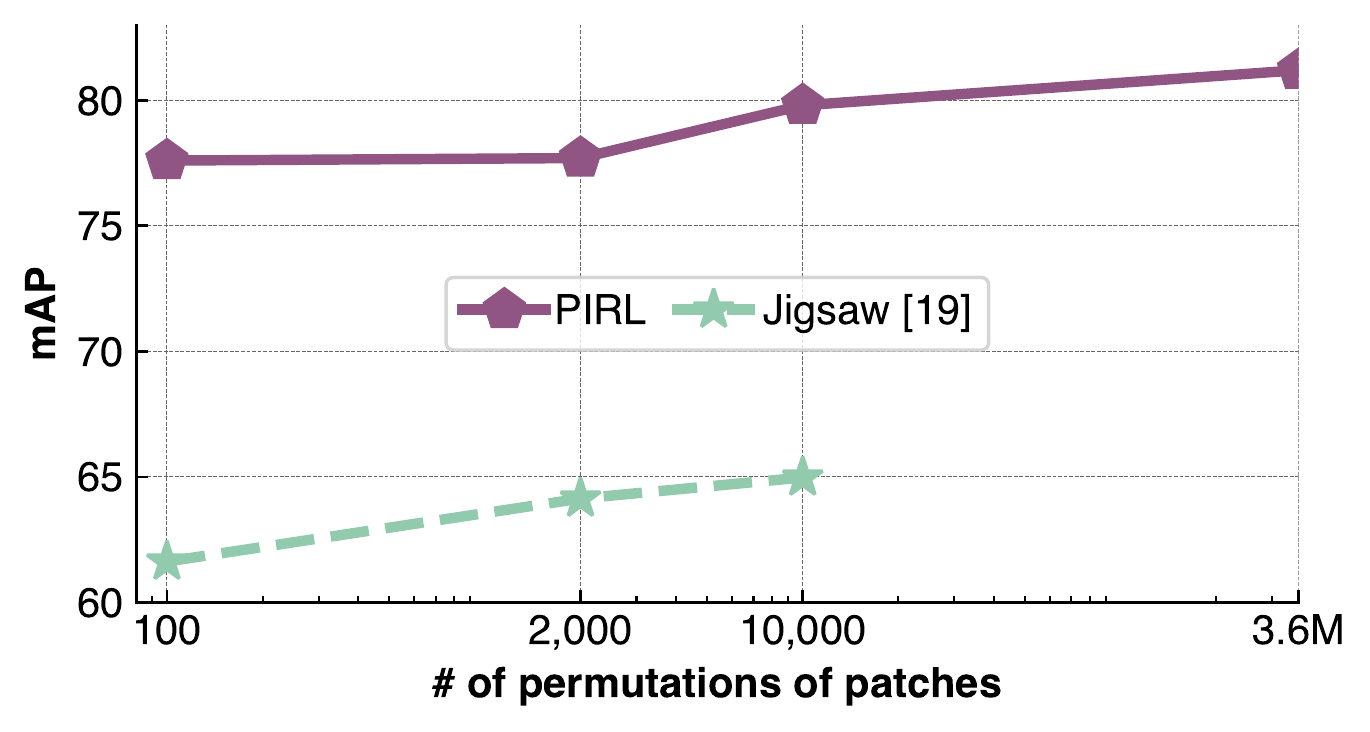}
\caption{\textbf{Effect of varying the number of patch permutations in $\calT$.}
Performance of linear image classification models trained on the \VOCseven dataset in terms of mAP. Models are initialized by \algorithmName and \jigsaw, varying the number of image transformations, $\calT$, from $1$ to $9! \approx 3.6$ million.} \label{fig:problem_complexity}
\end{figure}

\paragraph{What is the effect of the number of image transforms?}
Both in \algorithmName and \jigsaw, it is possible to vary the complexity of the task by varying the number of permutations of the nine image patches that are included in the set of image transformations, $\calT$. Prior work on \jigsaw suggests that increasing the number of possible patch permutations leads to better performance~\cite{noroozi2016unsupervised,goyal2019scaling}. However, the largest value $|\calT|$ can take is restricted because the number of learnable parameters in the output layer grows linearly with the number of patch permutations in models trained to solve the \jigsaw task. This problem does not apply to \algorithmName because it never outputs the patch permutations, and thus has a fixed number of model parameters.
As a result, \algorithmName can use all $9! \approx 3.6$ million permutations in $\calT$.

We study the quality of \algorithmName and \jigsaw as a function of the number of patch permutations included in $\calT$. To facilitate comparison with~\cite{goyal2019scaling}, we measure quality in terms of performance of linear models on image classification using the \VOCseven dataset, following the same setup as in Section~\ref{sec:linear_all}. The results of these experiments are presented in Figure~\ref{fig:problem_complexity}. The results show that \algorithmName outperforms \jigsaw for all cardinalities of $\calT$ but that \algorithmName particularly benefits from being able to use very large numbers of image transformations (\emph{i.e.}, large $| \calT |$) during training.

\begin{figure}
\centering
\includegraphics[width=\linewidth]{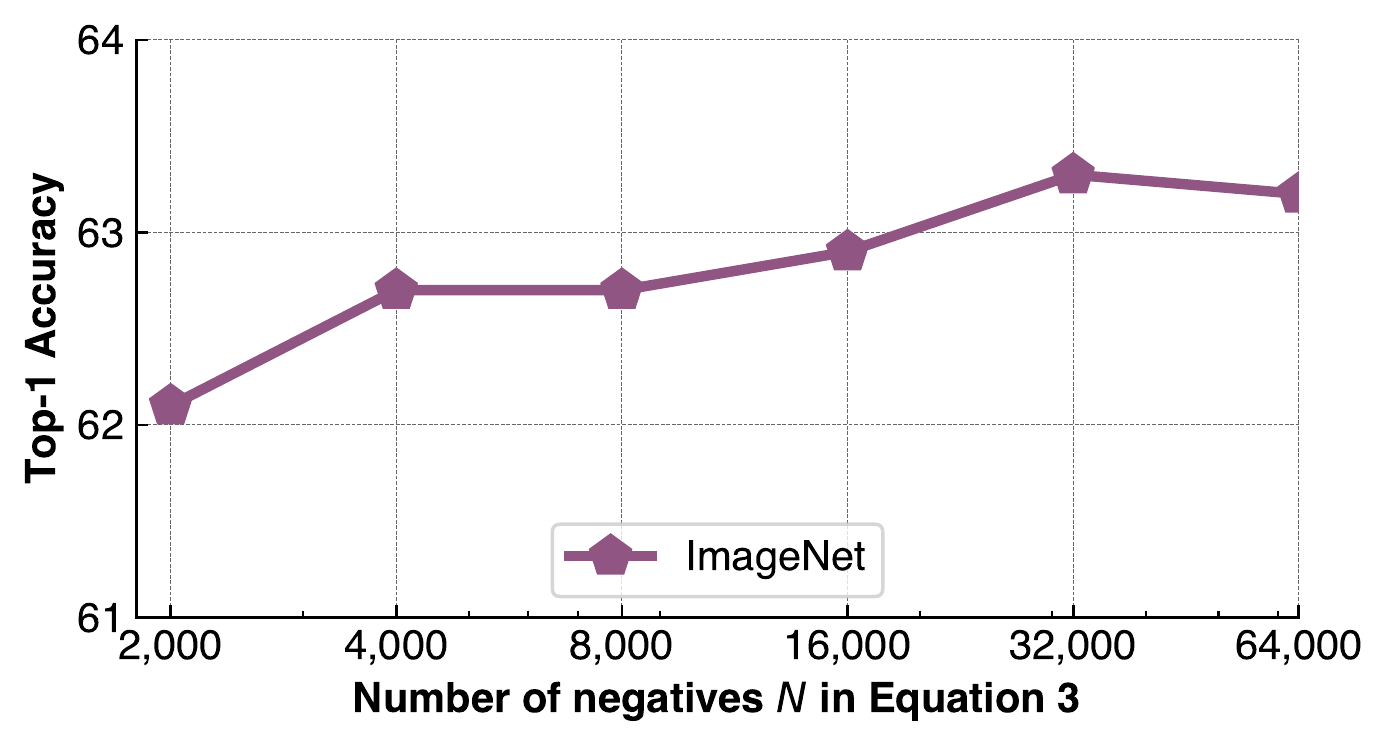}
\caption{\textbf{Effect of varying the number of negative samples.} Top-1 accuracy of linear classifiers trained to perform \ImNet classification using \algorithmName representations as a function of the number of negative samples, $N$.}
\label{fig:num_negs}
\end{figure}

\paragraph{What is the effect of the number of negative samples?}
We study the effect of the number of negative samples, $N$, on the quality of the learned image representations. We measure the accuracy of linear \ImNet classifiers on \underline{fixed} representations produced by \algorithmName as a function of the value of $N$ used in pre-training.  The results of these experiments are presented in Figure~\ref{fig:num_negs}. They suggest that increasing the number of negatives tends to have a positive influence on the quality of the image representations constructed by \algorithmName. 

\subsection{Generalizing \algorithmName to Other Pretext Tasks}
\label{sec:rotation}
Although we studied \algorithmName in the context of \jigsaw in this paper, \algorithmName can be used with any set of image transformations, $\calT$. We performed an experiment evaluating the performance of \algorithmName using the \rotation pretext task~\cite{gidaris2018unsupervised}. We define $\imageTransformSetNoMath$ to contain image  rotations by $\{0^\circ, 90^\circ, 180^\circ, 270^\circ\}$, and measure representation quality in terms of image-classification accuracy of linear models (see the supplemental material for details).

The results of these experiments are presented in Table~\ref{tab:pirl_rot} (top). In line with earlier results, models trained using \algorithmName (\rotation) outperform those trained using the \rotation pretext task of~\cite{gidaris2018unsupervised}. The performance gains obtained from learning a rotation-invariant representation are substantial, \eg $+11\%$ top-1 accuracy on \ImNet. We also note that \algorithmName (\rotation) outperforms NPID++ (see Table~\ref{tab:linear_all}). 
In a second set of experiments, we combined the pretext image transforms from both the \jigsaw and \rotation tasks in the set of image transformations, $\calT$. Specifically, we obtain $\bI^{t}$ by first applying a rotation and then performing a \jigsaw transformation. The results of these experiments are shown in Table~\ref{tab:pirl_rot} (bottom). The results demonstrate that combining image transforms from multiple pretext tasks can further improve image representations.

\section{Related Work}
\label{sec:related_work}

Our study is related to prior work that tries to learn characteristics of the image distribution without considering a corresponding (image-conditional) label distribution. A variety of work has studied reconstructing images from a small, intermediate representation, \eg, using sparse coding~\cite{olshausen1996},  adversarial training~\cite{donahue2017adversarial,mescheder2017unifying,donahue2019large}, autoencoders~\cite{masci2011stacked,ranzato2007unsupervised,vincent2008extracting}, or probabilistic versions thereof~\cite{salakhutdinov2009deep}. 

More recently, interest has shifted to specifying pretext tasks~\cite{doersch2015unsupervised} that require modeling a more limited set of properties of the data distribution. For video data, these pretext tasks learn representations by ordering video frames~\cite{misra2016shuffle,fernando2017self,opn,Wei18,ahsan2019video,xu2019self,kim2019self}, tracking~\cite{wang2015unsupervised,pathak2017learning}, or using cross-modal signals like audio~\cite{owens2016ambient,arandjelovic2018objects,arandjelovic2017look,bruno_avts,gao2018learning,owens}.

Our work focuses on image-based pretext tasks. Prior pretext tasks include image colorization~\cite{deshpande2015learning,iizuka2016let,larsson2016learning,larsson2017colorization,zhang2016colorful,zhang2017split}, orientation prediction~\cite{gidaris2018unsupervised}, affine transform prediction~\cite{zhang2019aet}, predicting contextual image patches~\cite{doersch2015unsupervised}, re-ordering image patches~\cite{goyal2019scaling,noroozi2016unsupervised,noroozi2018boosting,carlucci2019domain}, counting visual primitives~\cite{noroozi2017representation}, or their combinations~\cite{doersch2017multi}. In contrast, our works learns image representations that are invariant to the image transformations rather than covariant. 

\algorithmName is related to approaches that learn invariant image representations via contrastive learning~\cite{wu2018unsupervised,wang2015unsupervised,hjelm2018learning,dwibedi2019tcc,sermanet2018time}, clustering~\cite{caron2018deep,caron2019deep,noroozi2018boosting,wang2017transitive}, or maximizing mutual information~\cite{ji2019invariant,hjelm2018learning,bachman2019amdim}. \algorithmName is most similar to methods that learn representations that are invariant under standard data augmentation~\cite{hjelm2018learning,bachman2019amdim,wu2018unsupervised,ji2019invariant,dosovitskiy2016discriminative,xie2019unsupervised}. \algorithmName learns representations that are invariant to both the data augmentation and to the pretext image transformations. 
\begin{table}[!t]
\centering
\setlength{\tabcolsep}{0.2em}
\resizebox{\linewidth}{!}{

\begin{tabular}{l|c|acac}
\textbf{Method} & \textbf{Params} & \multicolumn{4}{c}{\textbf{Transfer Dataset}}\\
 & & {\small{\textbf{\ImNet}}} & {\small{\textbf{\VOCseven}}} & {\small{\textbf{\Placestwo}}} & {\small{\textbf{\iNat}}} \\
\thinline
\rotation~\cite{gidaris2018unsupervised} & 25.6M & 48.9 & 63.9 & 41.4 & 23.0\\

\algorithmShort (Rotation; \textbf{ours}) & 25.6M & 60.2 & 77.1 & 47.6 & 31.2 \\
$\mathbf{\Delta}$ of \algorithmShort & - &  \textbf{+11.3} &  \textbf{+13.2} &  \textbf{+6.2} &  \textbf{+8.2} \\
\thinline
\multicolumn{6}{l}{Combining pretext tasks using \algorithmName} \\
\thinline
\algorithmShort (Jigsaw; \textbf{ours}) & 25.6M & 62.2 &  79.8 & 48.5 & 31.2\\
\algorithmShort \small{(Rotation + Jigsaw; \textbf{ours})} & 25.6M & \textbf{63.1} & \textbf{80.3} & \textbf{49.7} & \textbf{33.6} \\
\thinline
\end{tabular}}
\vspace{-0.1in}
\caption{\textbf{Using \algorithmName with (combinations of) different pretext tasks.} Top-1 accuracy / mAP of linear image classifiers trained on \algorithmName image representations. \underline{Top}: Performance of \algorithmName used in combination with the \rotation pretext task~\cite{gidaris2018unsupervised}. \underline{Bottom:} Performance of \algorithmName using a combination of multiple pretext tasks.}

\label{tab:pirl_rot}
\end{table}

Finally, \algorithmName is also related to approaches that use a contrastive loss~\cite{hadsell2006dimensionality} in predictive learning~\cite{he2019moco,henaff2019data,oord2018representation,tian2019contrastive,han2019video,sun2019contrastive}. These prior approaches predict missing parts of the data, \eg, future frames in videos~\cite{oord2018representation,han2019video}, or operate on multiple views~\cite{tian2019contrastive}. In contrast to those approaches, \algorithmName learns invariances rather than predicting missing data.

\section{Discussion and Conclusion}
\label{sec:discussion}

We studied Pretext-Invariant Representation Learning (\algorithmName) for learning representations that are invariant to image transformations applied in self-supervised pretext tasks. The rationale behind \algorithmName is that invariance to image transformations maintains semantic information in the representation.
We obtain state-of-the-art results on multiple benchmarks for self-supervised learning in image classification and object detection. \algorithmName even outperforms supervised \ImNet pre-training on object detection. 

In this paper, we used \algorithmName with the \jigsaw and \rotation image transformations. In future work, we aim to extend to richer sets of transformations. We also plan to investigate combinations of \algorithmName with clustering-based approaches~\cite{caron2018deep,caron2019deep}. Like \algorithmName, those approaches use inter-image statistics but they do so in a different way. A combination of the two approaches may lead to even better image representations.

\vspace{-0.1in}
{\small
\paragraph{Acknowledgments:} We are grateful to Rob Fergus, and Andrea Vedaldi for encouragement and their feedback on early versions of the manuscript; Yann LeCun for helpful discussions; Aaron Adcock, Naman Goyal, Priya Goyal, and Myle Ott for their help with the code development for this research; and Rohit Girdhar, and Ross Girshick for feedback on the manuscript. We thank Yuxin Wu, and Kaiming He for help with \detectrontwo.}

{\small
\bibliographystyle{ieee_fullname}
\bibliography{refs}
}
\clearpage
\appendix
\section*{Supplemental Material}

\section{Training architecture and \\ Hyperparameters}

\paragraph{Base Architecture for \algorithmName} \algorithmName models in Section 3, 4 and 5 in the main paper are standard \resnetfifty~\cite{he2016deep} models with 25.6 million parameters. Figure 2 and Section 3.2 also use a larger model `\algorithmName-c2x' which has double the number of channels in each `stage' of the ResNet, \eg, 128 channels in the first convolutional layer, 4096 channels in the final \resfive~layer \etc, and a total of 98 million parameters. The heads $f(\cdot)$ and $g(\cdot)$ are linear layers that are used for the pre-training stage (Figure 3). When evaluating the model using transfer learning (Section 4 and 5 of the main paper), the heads $f$, $g$ are removed.

\paragraph{Training Hyperparameters for Section 3} \algorithmName models in Section 3 are trained for $800$ epochs using the ImageNet training set (1.28 million images). We use a batchsize of 32 per GPU and use a total of 32 GPUs to train each model. We optimize the models using mini-batch SGD with the cosine learning rate decay~\cite{loshchilov2016sgdr} scheme with an initial learning rate of $1.2\times10^{-1}$ and a final learning rate of $1.2\times10^{-4}$, momentum of $0.9$ and a weight decay of $10^{-4}$. We use a total of $32,000$ negatives to compute the NCE loss (Equation~\ref{eq:final_loss}) and the negatives are sampled randomly for each data point in the batch.

\paragraph{Training Hyperparameters for Section 4} The hyperparameters used are exactly the same as Section 3, except that the models are trained with $4096$ negatives and for $400$ epochs only. This results in a lower absolute performance, but the main focus of Section 4 is to analyze \algorithmName.

\paragraph{Common Data Pre-processing} We use data augmentations as implemented in PyTorch and the Python Imaging Library. These data augmentations are used for all methods including \algorithmName and NPID++. We do not use supervised policies like Fast AutoAugment. Our `geometric' data pre-processing involves extracting a randomly resized crop from the image, and horizontally flipping it. We follow this by `photometric' pre-processing that alter the RGB color values by using random values of color jitter, contrast, brightness, saturation, sharpness, equalize \etc transforms to the image.

\subsection{Details for \algorithmName with \jigsaw}
\paragraph{Data pre-processing} We base our \jigsaw implementation on~\cite{goyal2019scaling,kolesnikov2019revisiting}. To construct $\bI^{t}$, we extract a random resized crop that occupies at least 60\% of the image. We resize this crop to $255\times255$, and then divide into a $3\times3$ grid. We extract a patch of $64\times64$ from each of these grids and get $9$ patches. We apply photometric data augmentation (random color jitter, hue, contrast, saturation) to each patch independently and finally obtain the $9$ patches that constitute $\bI^{t}$.
\par \noindent The image $\bI$ is a $224\times224$ image obtained by applying standard data augmentation (flip, random resized crop, color jitter, hue, contrast, saturation) to the image from the dataset.

\paragraph{Architecture} The 9 patches of $\bI^{t}$ are individually input to the network to get their \resfive\ features which are average pooled to get a single $2048$ dimensional vector for each patch. We then apply a linear projection to these features to get a $128$ dimensional vector for each patch. These patch features are concatenated to get a $1152$ dimensional vector which is then input to another single linear layer to get the final $128$ dimensional feature $\bv_{\bI^{t}}$.
\par \noindent We feed forward the image $\bI$ and average pool its \resfive\ feature to get a $2048$ dimensional vector for the image. We then apply a single linear projection to get a $128$ dimensional feature $\bv_\bI$.

\subsection{Details for \algorithmName with \rotation}
\paragraph{Data pre-processing} We base our \rotation implementation on~\cite{gidaris2018unsupervised,kolesnikov2019revisiting}. To construct $\bI^{t}$ we use the standard data augmentation described earlier (geometric + photometric) to get a $224\times224$ image, followed by a random rotation from $\{0^{\circ}, 90^{\circ}, 180^{\circ}, 270^{\circ}\}$.

\par \noindent The image $\bI$ is a $224\times224$ image obtained by applying standard data augmentation (flip, random resized crop, color jitter, hue, contrast, saturation) to the image from the dataset.

\paragraph{Architecture} We feed forward the image $\bI^{t}$, average pool the \resfive\ feature, and apply a single linear layer to get the final $128$ dimensional feature for $\bv_{\bI^{t}}$.
\par \noindent We feed forward the image $\bI$ and average pool its \resfive\ feature to get a $2048$ dimensional vector for the image. We then apply a single linear projection to get a $128$ dimensional feature $\bv_\bI$.

\section{Object Detection}
\subsection{\VOCseven train+val set for detection}
In Section 3.1 of the main paper, we presented object detection results using the \VOCseventwelve training set which has 16K images. In this section, we use the smaller \VOCseven train+val set (5K images) for finetuning the Faster R-CNN C4 detection models. All models are finetuned using the Detectron2~\cite{wu2019detectron2} codebase and hyperparameters from~\cite{goyal2019scaling}. We report the detection AP in Table~\ref{tab:detection_voc07_train}. We see that the \algorithmName model outperforms the ImageNet supervised pretrained models on the stricter AP$^{75}$ and AP$^{\mathrm{all}}$ metrics without extra pretraining data or changes to the network architecture.
\par \noindent \textbf{Detection hyperparameters:} We use a batchsize of 2 images per GPU, a total of 8 GPUs and finetune models for $12.5$K iterations with the learning rate dropped by $0.1$ at $9.5$K iterations. The supervised and \jigsaw baseline models use an initial learning rate of $0.02$ with a linear warmup for $100$ iterations with a slope of $1/3$, while the NPID++ and \algorithmName models use an initial learning rate of $0.003$ with a linear warmup for $1000$ iterations and a slope of $1/3$. We keep the BatchNorm parameters of all models fixed during the detection finetuning stage.

\begin{table}[!h]
\centering
\setlength{\tabcolsep}{0.2em}\resizebox{\linewidth}{!}{
\begin{tabular}{l|c|ccc|r}
\textbf{Method} & \textbf{Network} & $\boldsymbol{\mathrm{AP}^\mathrm{all}}$ & $\boldsymbol{\mathrm{AP}^{50}}$ & $\boldsymbol{\mathrm{AP}^{75}}$ & $\boldsymbol{\Delta \mathrm{AP}^{75}}$ \\
\shline

Supervised & \resnetfiftyShort & 43.8 & 74.5 & 45.9 & =0.0\\
\thinline
\rowcolor{highlightRowColor} \jigsaw~\cite{goyal2019scaling} & \resnetfiftyShort & 37.7 & 64.9 & 38.0 & -7.9\\
\algorithmShort \textbf{(ours)} & \resnetfiftyShort & 44.7 & 73.4 & 47.0 & \textbf{+1.1}\\
\thinline
\rowcolor{highlightRowColor} NPID~\cite{wu2018unsupervised} & \resnetfiftyShort & -- & ~65.4$^*$ & -- \\
LA~\cite{zhuang2019local} & \resnetfiftyShort & -- & ~69.1$^*$ & -- \\
\rowcolor{highlightRowColor} Multi-task~\cite{doersch2017multi} & \resnethundredShort & -- & ~70.5$^*$ & -- \\
\thinline

\end{tabular}}

\caption{\textbf{Object detection on \VOCseven using Faster R-CNN.} Detection AP on the \VOCseven test set after finetuning Faster R-CNN models with a \resnetfifty backbone pre-trained using self-supervised learning on \ImNet. Results for supervised \ImNet pre-training are presented for reference. All methods are finetuned using the images from the \VOCseven train+val set. Numbers with $^*$ are adopted from the corresponding papers. We see that \algorithmName outperforms supervised pre-training without extra pre-training data or changes in the network architecture.}
\label{tab:detection_voc07_train}
\end{table}

\subsection{\VOCseventwelve train set for detection}
In Table~\ref{tab:detection_voc07}, we use the \VOCseventwelve training split and the \VOCseven test set as used in prior work~\cite{oord2018representation,goyal2019scaling}.
\par \noindent \textbf{Detection hyperparamters:} We use a batchsize of 2 images per GPU, a total of 8 GPUs and finetune models for $25$K iterations with the learning rate dropped by $0.1$ at $17$K iterations. The supervised and \jigsaw baseline models use an initial learning rate of $0.02$ with a linear warmup for $100$ iterations with a slope of $1/3$, while the NPID++ and \algorithmName models use an initial learning rate of $0.003$ with a linear warmup for $1000$ iterations and a slope of $1/3$. We keep the BatchNorm parameters of all models fixed during the detection finetuning stage.

\section{Linear Models for Transfer}
We train linear models on representations from the intermediate layers of a \resnetfifty model following the procedure outlined in~\cite{goyal2019scaling}. We briefly outline the hyperparameters from their work. The features from each of the layers are average pooled such that they are of about $9,000$ dimensions each. The linear model is trained with mini-batch SGD using a learning rate of $0.01$ decayed by $0.1$ at two equally spaced intervals, momentum of $0.9$ and weight decay of $5\times10^{-4}$. We train the linear models for $28$ epochs on \ImNet (1.28M training images), $14$ epochs on \Placestwo (2.4M training images) and for $84$ epochs on iNaturalist-2018 (437K training images). Thus, the number of parameter updates for training the linear models are roughly constant across all these datasets. We report the center crop top-1 accuracy for the \ImNet, \Placestwo and iNaturalist-2018 datasets in Table~\ref{tab:linear_all}.

For training linear models on \VOCseven, we train linear SVMs following the procedure in~\cite{goyal2019scaling} and report mean average Precision (mAP).

\begin{table*}[!t]
\centering
\setlength{\tabcolsep}{0.65em}\resizebox{\linewidth}{!}{
\begin{tabular}{l|ccccccccccc}
& \multicolumn{5}{c}{\textbf{\ImNet}} & \multicolumn{5}{c}{\textbf{\Placestwo}} \\
 \textbf{Method} & \textbf{\mytexttt{conv1}} & \textbf{\mytexttt{res2}} & \textbf{\mytexttt{res3}} & \textbf{\mytexttt{res4}} & \textbf{\mytexttt{res5}} & \textbf{ } & \textbf{\mytexttt{conv1}} & \textbf{\mytexttt{res2}} & \textbf{\mytexttt{res3}} & \textbf{\mytexttt{res4}} & \textbf{\mytexttt{res5}}\\
\shline
Random & 9.6 & 13.7 & 12.0 & 8.0 & 5.6 &   & 12.9 & 16.6 & 15.5 & 11.6 & 9.0\\
\rowcolor{highlightRowColor} \ImNet Supervised & 11.6 & 33.3 & 48.7 & 67.9 & 75.5 &   & 14.8 & 32.6 & 42.1 & 50.8 & 51.5\\
\Placestwo Supervised & 13.2 & 31.7 & 46.0 & 58.2 & 51.7 &   & 16.7 & 32.3 & 43.2 & 54.7 & 62.3\\

\thinline
\rowcolor{highlightRowColor} \jigsaw~\cite{goyal2019scaling} & 12.4 & 28.0 & 39.9 & 45.7 & 34.2 &   & 15.1 & 28.8 & 36.8 & 41.2 & 34.4\\
Coloriz.~\cite{goyal2019scaling} & 10.2 & 24.1 & 31.4 & 39.6 & 35.2 &   & 18.1 & 28.4 & 30.2 & 31.3 & 30.4\\

\rowcolor{highlightRowColor} NPID++~\cite{wu2018unsupervised} &  6.6 & 27.4 & 38.5 & 53.1 & 59 && 9.7 & 26.2 & 34.9 & 43.7 & 46.4\\

\algorithmName \textbf{(Ours)} & 6.2 & 30 & 40.1 & 56.6 & 63.6 && 8.9 & 29 & 35.8 & 45.3 & 49.8 \\

\thinline
\multicolumn{12}{c}{Different Evaluation Setup}\\
\thinline
\rowcolor{highlightRowColor} Kolesnikov \etal~\cite{kolesnikov2019revisiting}& -- & -- & -- & 47.7 & -- &   & -- & -- & -- & -- & --\\
NPID~\cite{wu2018unsupervised} & 15.3 & 18.8 & 24.9 & 40.6 & 54.0 &   & 18.1 & 22.3 & 29.7 & 42.1 & 45.5\\

\shline
\end{tabular}}
\caption{\textbf{\resnetfifty linear evaluation on \ImNet and \Placestwo datasets:} We report the performance of all the layers following the evaluation protocol in~\cite{goyal2019scaling}. All numbers except NPID++ and Ours are from their respective papers.}
\label{tab:rn50_linear_per_layer}
\end{table*}
\paragraph{Per layer results}
 The results of all these models are in Table~\ref{tab:rn50_linear_per_layer}.

\end{document}